\documentclass[11pt]{article}
\usepackage[preprint]{acl}

\usepackage{times}
\usepackage{latexsym}
\usepackage[T1]{fontenc}
\usepackage[utf8]{inputenc}
\usepackage{microtype}
\usepackage{inconsolata}

\usepackage{graphicx}

\usepackage{amsmath}
\usepackage{amssymb}

\usepackage{booktabs}
\usepackage{multirow}
\usepackage{array}

\usepackage{placeins}
\usepackage{float}
\usepackage{xspace}

\title{TextFake: Benchmarking AI-Generated Image Detection on Text-Rich Images}

\author{
  \textbf{Yuning Zhang\textsuperscript{1,2}},
  \textbf{Changtao Miao\textsuperscript{3,*}},
  \textbf{Mingyu Liao\textsuperscript{1,2}},
  \textbf{Tingyu Liu\textsuperscript{1,2}},
  \textbf{Xinghao Wang\textsuperscript{1,2}}
\\
  \textbf{Tao Gong\textsuperscript{1,2,$\dagger$}},
  \textbf{Qi Chu\textsuperscript{1,2}},
  \textbf{Nenghai Yu\textsuperscript{1,2}}
\\[4pt]
  \textsuperscript{1}School of Cyber Science and Technology, University of Science and Technology of China \\
  \textsuperscript{2}Anhui Province Key Laboratory of Digital Security \quad
  \textsuperscript{3}Individual Researcher
}

\begin{document}
\maketitle
{
  \let\thefootnote\relax
  \footnotetext{$^*$ Project Lead. $^\dagger$ Corresponding Author.}
}

\begin{abstract}Recent AI-generated image (AIGI) detectors perform well on natural-image benchmarks, but their behavior on text-rich forgeries, such as fabricated screenshots, documents, and news pages prevalent in misinformation, remains untested. We introduce \textbf{TextFake}, a 20,000-image benchmark for text-rich AIGI detection spanning 28 languages, 4 topic categories, and 2 scene modalities. Fake images are synthesized via a four-stage pipeline that annotates real images along three controlled dimensions and generates counterparts through distribution-aligned structured prompting, ruling out covariate shortcuts. Zero-shot evaluation of 14 specialized detectors and 3 frontier VLM APIs reveals a large systematic gap: no method exceeds 80\% accuracy, with some dropping over 60\% from natural-image benchmarks. Diagnostic evaluations identify three failure modes: the \textit{Text Density Curse}, where dense glyphs overwhelm low-level detectors; \textit{Cloaking via Rendering Fidelity}, where stronger text rendering suppresses generative artifacts; and \textit{Threshold Collapse}, where routine perturbations drive detectors toward chance-level performance.
\end{abstract}

\section{Introduction}
\label{sec:intro}

\begin{figure}[t]
\centering
\includegraphics[width=\columnwidth]{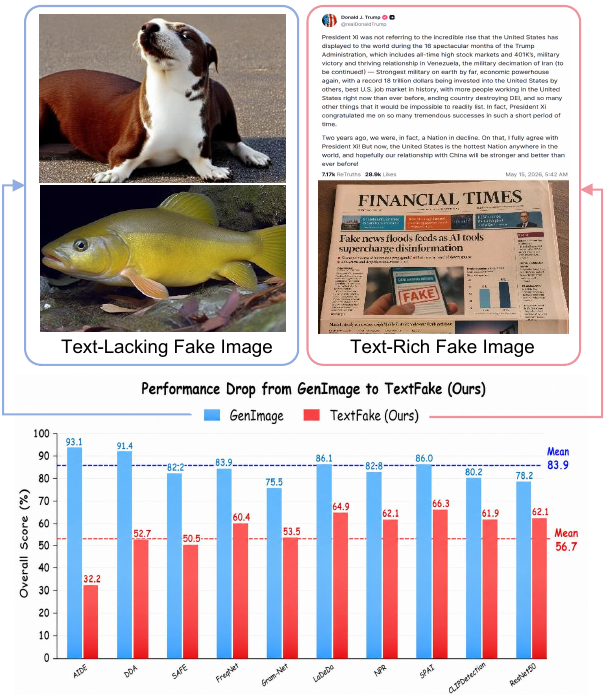}
\caption{Text-rich fake images expose a critical detection gap: across representative detectors, mean accuracy drops from 86.7\% on GenImage to 51.8\% on TextFake, a decline of 34.9 percentage points.}
\label{fig:teaser}
\end{figure}
Despite rapid progress in AI-generated image (AIGI) detection~\cite{intro1}, existing benchmarks remain dominated by natural-scene images, leaving text-rich forgeries largely unexplored.
Fabricated screenshots, fake news pages, and official documents are prevalent in misinformation and fraud~\cite{intro2}; as frontier generators grow capable of rendering multilingual text and realistic layouts, this gap is increasingly consequential.

Existing AIGI benchmarks have grown from early deepfake collections~\cite{uadfv,forgeryNet} to large natural-image evaluations such as CIFAKE~\cite{cifake} and GenImage~\cite{genimage}.
Yet text-rich imagery appears in none of them as a design dimension or evaluation target.
Because no existing benchmark covers this modality, detection methods have never been evaluated on text-rich forgeries and their behavior here is unknown~\cite{intro3,mirsky2021creation,sukhija2025document}, which leaves a task that is both high-stakes and unexplored.

We introduce \textbf{TextFake}, a 20,000-image benchmark for text-rich AIGI detection covering 28 languages, two scene modalities, and four topics, with real and synthetic subsets distributionally aligned to reduce dataset shortcuts.

Zero-shot evaluation of 14 specialized detectors and three frontier VLM APIs shows that no method exceeds 80\% accuracy, while several spatial- and frequency-based detectors~\cite{cnndetect,qian2020thinking} perform near chance. 
A four-stage pipeline reveals that performance declines as text density increases; higher-fidelity generators produce images that are harder to detect; and JPEG compression, Gaussian noise, and screenshot-induced moir\'{e} patterns cause severe prediction shifts. By aligning distributions across scene, topic, and language, the pipeline rules out covariate shortcuts and attributes these gaps to the intrinsic challenges posed by text-rich content.


Our contributions are:
\begin{itemize}
    \item We construct \textbf{TextFake}, a 20,000-image benchmark for text-rich AIGI detection spanning 28 languages, two scene modalities, and four topics, with real and synthetic subsets distributionally aligned.
    \item We introduce a four-stage pipeline that collects multilingual real images, annotates them via OCR and VLM classifiers, synthesizes fake counterparts via distribution-aligned structured prompting, and normalizes metadata for fair comparison.
    \item We evaluate 14 specialized detectors and three frontier VLM APIs on TextFake, showing that natural-scene benchmark performance does not transfer, and conduct four diagnostic analyses (text density, rendering fidelity, perturbation robustness, and language/scene bias) that surface concrete failure modes.
\end{itemize}

\section{Related Work}
\label{sec:related_work}

\paragraph{AIGI Detection Benchmarks.}
AI-generated image detection has progressed from early deepfake datasets~\cite{uadfv,forgeryNet} to open-domain natural-image benchmarks such as CIFAKE~\cite{cifake}, GenImage~\cite{genimage}, and cross-generator evaluations~\cite{artificial}. Specialized work has recently extended to scientific figures~\cite{scifigdetect}. Despite this progress, all mainstream benchmarks are dominated by natural scenes; text-rich modalities (fabricated screenshots, official documents, and news pages) remain unevaluated. These are socially consequential because a single altered number or name can change the evidential meaning of an image, and they directly stress-test the text-rendering limits of frontier generators. TextFake fills this gap.

\paragraph{Detection Paradigms.}
Current detectors broadly follow three lines. Spatial and frequency methods~\cite{resnet50,cnndetect,gramnet,npr,freenet,dffreq,safe,spai} extract hand-crafted or learned features from pixel statistics, edge patterns, and spectral coefficients to flag low-level artifacts left by generative pipelines, cues that text-rich images naturally mask with dense, legitimate high-frequency glyph structures. Foundation-model-based detectors~\cite{aide,clipdetection,ladeda,cospy,dda,gapl} instead probe or fine-tune large pretrained vision encoders (e.g., CLIP, DINOv2), relying on their broad visual priors to generalize across unseen generators. Proprietary VLM APIs~\cite{gemini3,gpt54,claudesonnet46} take this further by framing detection as a visual question-answering task, prompting multimodal LLMs in a zero-shot manner. The latter two paradigms offer richer semantics but are not attuned to pixel-level typographic authenticity.

\section{TextFake: A Distributionally Aligned Benchmark}
\label{sec:dataset}

Existing AIGI detection benchmarks mainly target open-domain natural images~\cite{genimage,cifake,forgeryNet}, leaving text-rich forgeries underexplored. This setting is important because screenshots, documents, and interface images contain dense glyph structures, layout regularities, and acquisition artifacts that differ substantially from natural scenes. To evaluate detector robustness in this high-risk regime, we construct \textbf{TextFake}, a multilingual benchmark with 20,000 text-rich images, evenly split into 10,000 real and 10,000 AI-generated samples.

TextFake is built around \textit{distributional control}. Each image is annotated along three dimensions: scene modality, semantic topic, and language. We align real and synthetic subsets on scene and topic distributions, while maintaining broad multilingual coverage for language-level analysis. This design reduces shortcut learning~\cite{textfake1} and supports both aggregate evaluation and failure diagnosis.

\begin{figure*}[t]
\centering
\includegraphics[width=\textwidth]{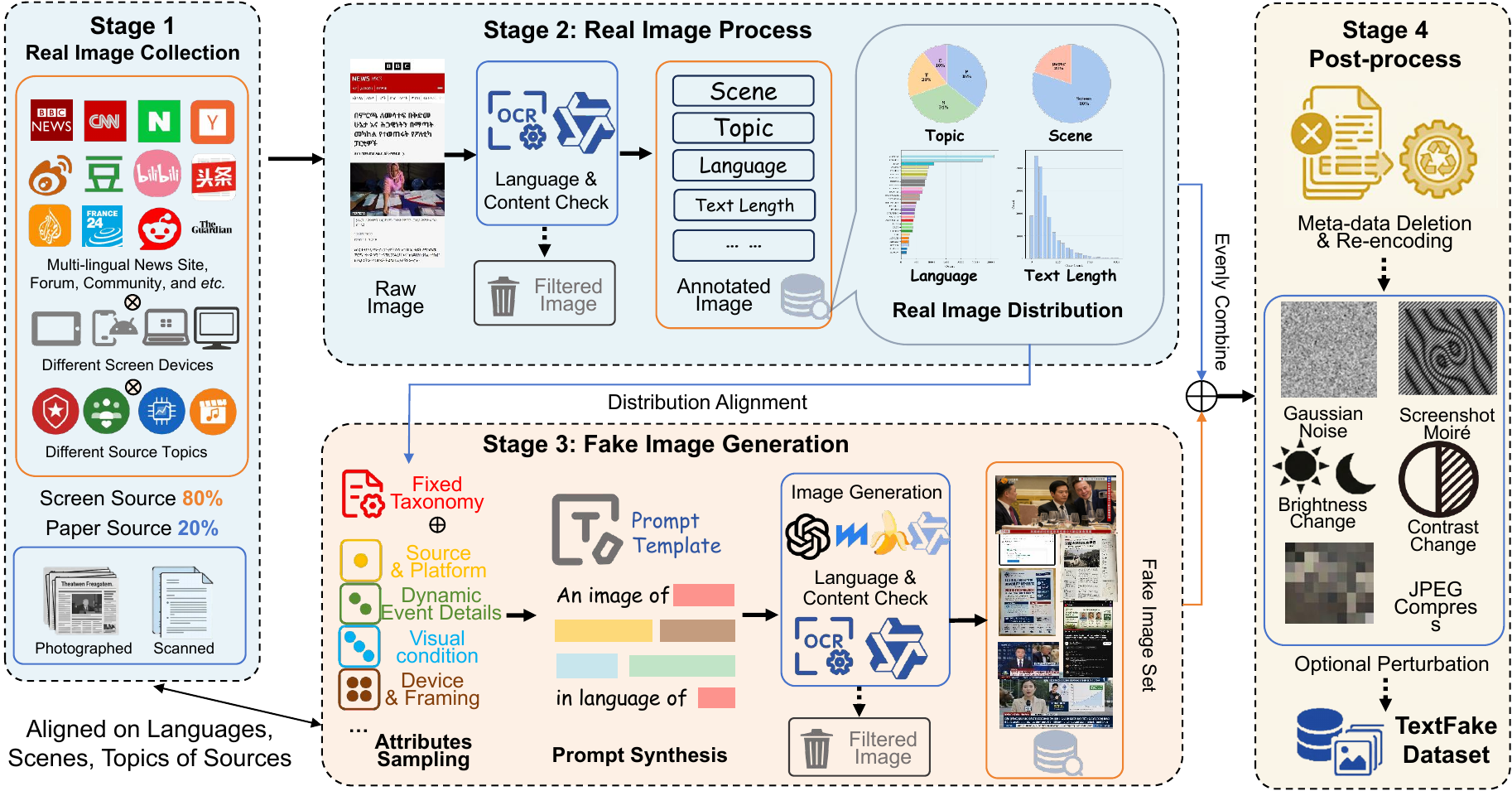}
\caption{Overview of the TextFake construction pipeline. \textbf{Stage~1} collects multilingual text-rich images from diverse sources including news sites, forums, and online communities, covering both screen captures and photographed or scanned paper documents. \textbf{Stage~2} applies OCR and VLM-based classifiers to annotate each image with scene type, topic category, language, and text length, filtering out samples that fail language or content checks, and producing a statistical profile of the real-image distribution. \textbf{Stage~3} generates fake counterparts through structured prompt synthesis, sampling attributes to align with the real-image distribution across languages, scenes, and topics. \textbf{Stage~4} evenly combines real and synthetic images, strips file metadata, uniformly re-encodes all samples to eliminate format-level shortcuts, and applies optional perturbations such as Gaussian noise, JPEG compression, and brightness changes for robustness evaluation.}
\label{fig:pipeline}
\end{figure*}

\subsection{Construction Pipeline}
\label{subsec:pipeline}

Figure~\ref{fig:pipeline} shows the four-stage construction pipeline. First, we collect real text-rich images from multilingual news sites, forums, and online communities, covering both screen captures and photographed or scanned paper documents. Second, each image is processed with OCR~\cite{ladeda} and VLM-based~\cite{achiam2023gpt} filters to obtain scene, topic, language, and text-length annotations. Third, synthetic images are generated through structured prompt synthesis. Each prompt combines fixed taxonomy attributes, source and platform information, dynamic event details, visual conditions, and device framing, with all attributes sampled to match the real-image distribution. Finally, real and synthetic images are combined, stripped of file metadata, uniformly JPEG re-encoded, and optionally perturbed for robustness evaluation.

\begin{figure}[!ht]
\centering
\fbox{
\begin{minipage}{0.94\columnwidth}
\small
\textbf{Structured prompt template}\\[0.5em]
\textit{
A [UI\_MODE] [SCENE] of [PLATFORM] in [LANGUAGE] about [TOPIC]: [DYNAMIC\_CONTENT].\\
UI layout: [UI\_ELEMENTS].\\
Aspect ratio: [ASPECT\_RATIO].\\
Resolution: [RESOLUTION].\\
Device: [DEVICE].\\
Capture condition: [PHYSICAL\_TRACE].\\
All text content is in [LANGUAGE] ([NATIVE\_SCRIPT]).
}
\end{minipage}
}
\caption{Structured prompt template used for synthetic image generation. Each prompt is assembled from a fixed taxonomy of attributes including topic category, language, and scene modality, combined with source and platform metadata, dynamically sampled event details, visual conditions such as lighting, resolution, and background clutter, and device framing cues such as screen bezels and paper edges. These elements are drawn to approximate the distributional characteristics of the corresponding real images, ensuring that the generated samples present realistic and diverse text-rich scenes rather than stylistically uniform outputs.}
\label{fig:prompt_template}
\end{figure}

\subsection{Controlled Dimensions}

TextFake controls three major dimensions. For scene modality, we use an 80:20 screen--paper prior in both real and fake subsets. Screen images contain digital rendering and anti-aliased glyph edges, while paper images include printing artifacts, paper texture, sensor noise, and moir\'{e} patterns.

For semantic topic, we use four classes: \textit{Politics \& Military}, \textit{Society \& Livelihood}, \textit{Technology \& Finance}, and \textit{Culture \& Entertainment}. The maximum real--fake topic gap is only 1.3 percentage points, reducing the risk that detectors exploit topic imbalance rather than authenticity cues.

For language, TextFake covers 28 languages from 12 language families, spanning both high-resource languages and typologically distinct scripts. This enables analysis of how glyph topology, stroke density, and script-specific frequency patterns affect detection.

\begin{table}[t]
    \centering
    \caption{Summary of TextFake dimensions and controls.}
    \label{tab:dataset_stats}
    \resizebox{0.96\columnwidth}{!}{
    \begin{tabular}{ll}
    \toprule
    \textbf{Dimension} & \textbf{Benchmark design} \\
    \midrule
    Total size & 20,000 images: 10,000 real and 10,000 fake \\
    Scene & 80:20 screen--paper prior in both real and fake subsets \\
    Topic & Four classes \\
    Language & 28 languages across 12 language families \\
    Real sources & 8,002 screenshots and 1,998 images of paper \\
    Quality control & VLM safety filtering + human spot-checking \\
    \bottomrule
    \end{tabular}
    }
\end{table}

\subsection{Dataset Summary and Quality Assurance}

Table~\ref{tab:dataset_stats} summarizes the benchmark. TextFake contains 20,000 images, including 10,000 real and 10,000 AI-generated samples. The real subset consists of 8,002 screenshots and 1,998 images of paper, while the fake subset is generated to match the benchmark design. All images pass language and content filtering, and VLM-assigned scene and topic labels are spot-checked by human annotators, with estimated agreement above 98\%. Metadata removal and uniform JPEG re-encoding are applied to all images before release.

\begin{figure}[t]
\centering
\includegraphics[width=\columnwidth]{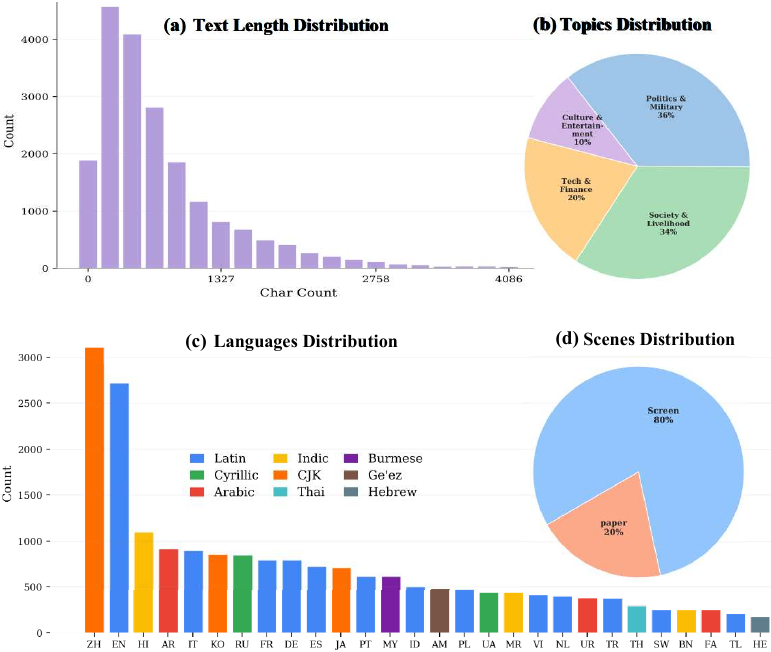}
\caption{Distribution of TextFake across the three controlled dimensions: scene modality (screen vs.\ paper), semantic topic, and language family. Real and synthetic subsets are aligned in scene and topic composition to minimize distributional shortcuts.}
\label{fig:dataset_stats}
\end{figure}
\section{Experiments: The Illusion of Generalization}
\label{sec:main_results}

We evaluate whether detectors that perform well on standard natural-image AIGI benchmarks, specifically GenImage~\cite{genimage}, remain reliable on text-rich images. The central question is not whether detectors can recognize synthetic natural scenes, but whether their learned forensic priors transfer to dense screenshots, interfaces, and documents.

\begin{table*}[t]
\centering
\caption{Detection performance (\%) on TextFake (Ours). Acc. = Overall Accuracy; Real Acc. = Accuracy on real image set; Fake Acc. = Accuracy on fake image set; F1 = macro F1; AP = average precision. AP is not applicable (N/A) for VLM APIs, which do not output probability scores.}
\label{tab:main_results}
\footnotesize
\setlength{\tabcolsep}{3pt}
\begin{tabular*}{\textwidth}{@{\extracolsep{\fill}}lccccc}
\toprule
\textbf{Method} & \textbf{Acc.} & \textbf{Real Acc.} & \textbf{Fake Acc.} & \textbf{F1} & \textbf{AP} \\
\midrule
ResNet50~\cite{resnet50}            & 62.1 & 98.8 & 25.4 & 40.2 & 76.1 \\
NPR~\cite{npr}                      & 62.1 & \textbf{99.5} & 24.7 & 39.5 & 73.8 \\
CNNDetection~\cite{cnndetect}       & 54.2 & 98.4 & 10.1 & 18.1 & 66.9 \\
Gram-Net~\cite{gramnet}             & 53.5 & 97.3 &  9.6 & 17.1 & 59.9 \\
DFFreq~\cite{dffreq}                & 66.0 & 99.6 & 32.5 & 48.9 & 73.2 \\
FreqNet~\cite{freenet}              & 60.4 & 94.1 & 26.8 & 40.3 & 67.5 \\
SAFE~\cite{safe}                    & 50.5 & 83.4 & 17.5 & 26.1 & 52.2 \\
SPAI~\cite{spai}                    & 66.3 & 87.3 & 45.4 & 57.4 & 70.3 \\
CLIPDetection~\cite{clipdetection}  & 61.9 & 71.8 & 52.0 & 57.7 & 65.6 \\
LaDeDa~\cite{ladeda}                & 64.9 & 99.3 & 30.5 & 46.5 & 72.5 \\
AIDE~\cite{aide}                    & 32.2 & 57.7 &  6.8 &  9.1 & 35.3 \\
CO-SPY~\cite{cospy}                 & 59.1 & 89.9 & 28.3 & 40.9 & 68.2 \\
DDA~\cite{dda}                      & 52.8 & 94.0 & 11.6 & 19.7 & 54.2 \\
GAPL~\cite{gapl}                    & \textbf{79.3} & 82.5 & \textbf{76.2} & \textbf{78.7} & \textbf{88.3} \\
Gemini-3-Pro-Preview~\cite{gemini3}       & 77.9 & 93.4 & 55.1 & 66.8 & N/A \\
GPT-5.4~\cite{gpt54}                      & 67.6 & 97.4 & 37.8 & 53.8 & N/A \\
Claude-Sonnet-4.6~\cite{claudesonnet46}   & 74.3 & 98.8 & 50.9 & 67.0 & N/A \\
\bottomrule
\end{tabular*}
\end{table*}

\subsection{Experimental Setup}

\textbf{Benchmark and preprocessing.}
All evaluations are conducted in a zero-shot setting~\cite{clipdetection}. To prevent detectors from exploiting source metadata, such as file extensions or initial compression signatures, every image is uniformly re-encoded as JPEG at 99\% quality before evaluation~\cite{cnndetect}. The primary metric is overall classification accuracy. For diagnostic analyses, we additionally report real accuracy and fake accuracy to identify class-specific threshold shifts.

\textbf{Evaluated baselines.}
We benchmark 18 evaluation configurations derived from 17 detection methods~\cite{artificial}. The baselines are grouped into three paradigms:
\begin{enumerate}
    \item[(1)] \textit{Spatial and frequency methods}, which target local pixel anomalies, gradient artifacts, or high-frequency spectral traces. This group includes ResNet50~\cite{resnet50}, CNNDetection~\cite{cnndetect}, NPR~\cite{npr}, FreqNet~\cite{freenet}, and additional benchmarked forensic detectors such as Gram-Net~\cite{gramnet}, DFFreq~\cite{dffreq}, SAFE~\cite{safe}, and SPAI~\cite{spai}.
    \item[(2)] \textit{Foundation-model-based methods}, which use representations from large-scale pretrained vision or vision-language models. This group includes AIDE~\cite{aide}, CLIPDetection~\cite{clipdetection}, LaDeDa~\cite{ladeda}, DDA~\cite{dda}, GAPL~\cite{gapl}, and CO-SPY~\cite{cospy}. For CO-SPY, we evaluate both SD-v1.4 and ProGAN initialization variants.
    \item[(3)] \textit{Proprietary VLM APIs}: Gemini-3-Pro-Preview~\cite{gemini3}, GPT-5.4~\cite{gpt54}, and Claude-Sonnet-4.6~\cite{claudesonnet46}, evaluated through direct zero-shot visual prompting.
\end{enumerate}

\subsection{Results and Analysis}
Table~\ref{tab:main_results} reports per-class accuracy, macro F1, and average precision for all evaluated methods on TextFake. No method exceeds 80\% overall accuracy, and the results reveal a consistent conservative bias across paradigms.

\textbf{Degradation of conventional artifact priors.}
Spatial and frequency methods exhibit a systematic pattern: high Real Acc.\ paired with very low Fake Acc. DFFreq reaches 99.6\% Real Acc.\ but only 32.5\% Fake Acc.; ResNet50 pairs 98.8\% Real Acc.\ with 25.4\% Fake Acc.; SAFE and FreqNet both fall below 20\% Fake Acc. These methods, calibrated on natural-image benchmarks, default to predicting authenticity on text-rich content, where dense glyph structures suppress the spectral anomalies they rely on. AP scores in the range 52--76 confirm that the underlying discriminative signal is weak rather than merely misthresholded.

Foundation-model-based detectors show similar or more extreme imbalance. AIDE achieves only 32.2\% overall Acc., with Fake Acc.\ of just 6.8\% and AP of 35.3. DDA pairs 94.0\% Real Acc.\ with 11.6\% Fake Acc. The notable exception is GAPL, which achieves 79.3\% Acc., the highest among all evaluated methods, with a roughly balanced per-class profile (Real Acc.\ 82.5\%, Fake Acc.\ 76.2\%, F1 78.7, AP 88.3), suggesting that its foundation-model representations are less sensitive to the low-level frequency shifts introduced by dense text.

\textbf{Performance limits of frontier VLMs.}
VLM APIs follow the same conservative pattern. GPT-5.4 pairs 97.4\% Real Acc.\ with only 37.8\% Fake Acc.\ (67.6\% overall); Claude-Sonnet-4.6 achieves 98.8\% Real Acc.\ but only 50.9\% Fake Acc.\ (74.3\% overall). Gemini-3-Pro-Preview produces the most balanced VLM profile (Real Acc.\ 93.4\%, Fake Acc.\ 55.1\%, 77.9\% overall). Across all three, semantic reasoning alone does not provide reliable forensic access to the low-level cues embedded in high-fidelity text-rich imagery.

Within TextFake, accuracy also varies across the four generator subsets: SPAI obtains 89.8\% on Qwen-image2 but drops to 72.4\% on Seedream5l, while DFFreq ranges from 79.2\% (Nano2) to 50.2\% (Seedream5l). This within-benchmark variation indicates that text-rendering fidelity affects forensic separability even within the text-rich domain, motivating the analysis in Section~\ref{subsec:cloaking}.

\section{Diagnostic Evaluation: Decoding the "Invisible Ink"}
\label{sec:diagnostics}

Aggregate accuracy alone does not explain why detectors fail in the text-rich regime. We introduce four diagnostic evaluations isolating: text density, text-rendering fidelity, transmission perturbations, and language and scene bias.

\subsection{The Text Density Curse}
\label{subsec:density_curse}

\paragraph{Setup \& Metric.}
We hypothesize that the volume of visible text disrupts conventional artifact extraction by introducing dense, legitimate high-frequency structure~\cite{watch,cnndetect}. To test this hypothesis, we run OCR on each image and use the recognized character count as a proxy for text density~\cite{ladeda}. Images are stratified into three percentile bins: LOW (0--10\%), MID (45--55\%), and HIGH (90--100\%). We report $\Delta(H-L)$ as the absolute accuracy change from the LOW bin to the HIGH bin.

\begin{table}[t]
\centering
\caption{Accuracy (\%) by text-density bin for representative detectors.}
\label{tab:density_bins}
\resizebox{\columnwidth}{!}{
\begin{tabular}{lcccc}
\toprule
\textbf{Method} & \textbf{LOW} & \textbf{MID} & \textbf{HIGH} & $\boldsymbol{\Delta}$ \\
\midrule
ResNet50~\cite{resnet50} & 91.8 & 55.8 & 54.5 & -37.3 \\
CNNDet~\cite{cnndetect} & 89.1 & 46.6 & 48.1 & -41.1 \\
Gram-Net~\cite{gramnet} & 88.6 & 45.2 & 45.6 & -43.0 \\
DFFreq~\cite{dffreq} & 94.3 & 59.4 & 56.5 & -37.8 \\
SPAI~\cite{spai} & 81.9 & 62.3 & 59.0 & -22.9 \\
GAPL~\cite{gapl}    & 77.0 & 78.9 & 80.8 & +3.8 \\
Gemini-3P~\cite{gemini3} & 86.8 & 75.7 & 76.8 & -10.0 \\
Sonnet-4.6~\cite{claudesonnet46} & 92.6 & 75.2 & 81.3 & -11.3 \\
GPT-5.4~\cite{gpt54}   & 90.5 & 62.9 & 72.5 & -18.0 \\
\bottomrule
\end{tabular}
}
\end{table}

\begin{figure}[t]
\centering
\includegraphics[width=\columnwidth]{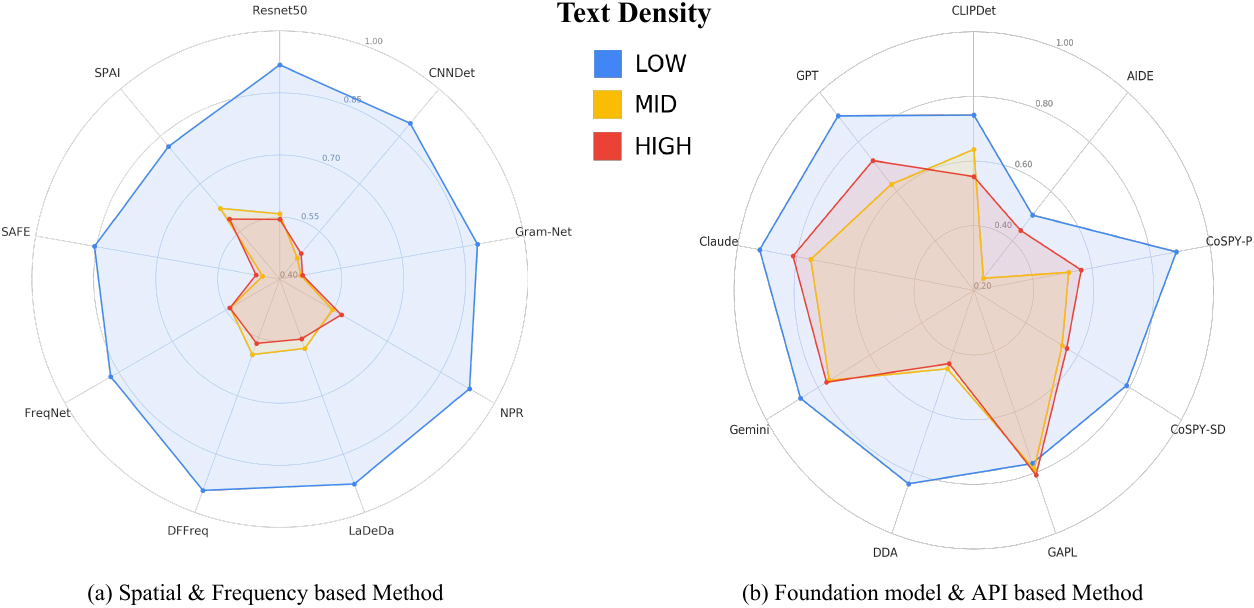}
\caption{Accuracy trends over character-count percentiles for \textbf{(a)} spatial and frequency methods and \textbf{(b)} foundation model and VLM API methods (VLMs shown as dashed lines). Highlighted regions mark the LOW (blue), MID (yellow), and HIGH (red) density bins. Spatial/frequency methods decline steeply from the LOW region, while foundation models and VLM APIs exhibit smaller but consistent degradation.}
\label{fig:text_density_trend}
\end{figure}

\begin{figure}[t]
\centering
\includegraphics[width=\columnwidth]{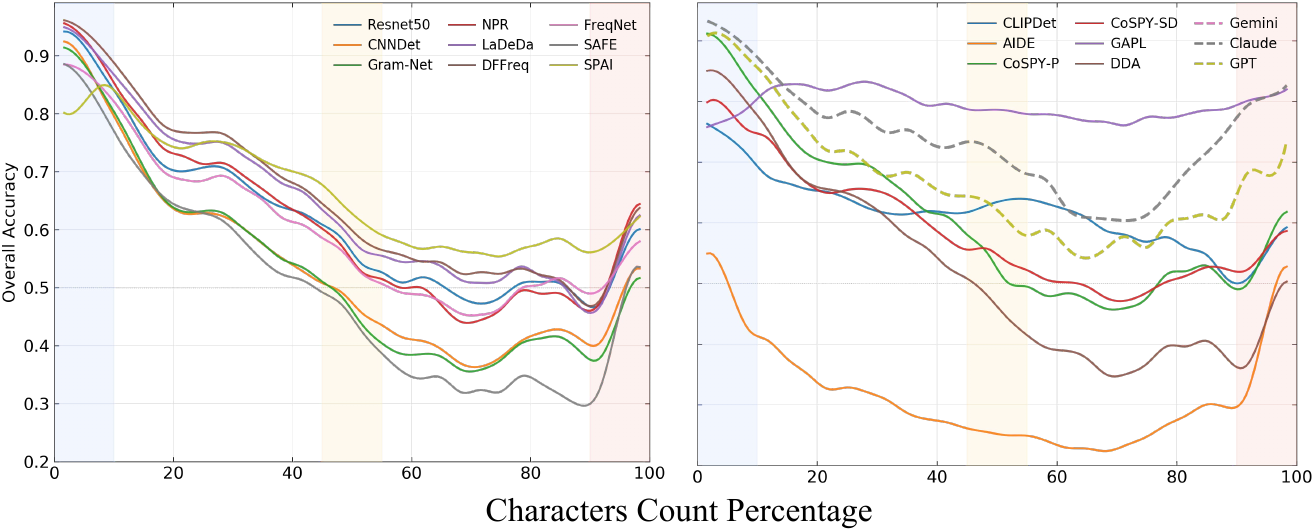}
\caption{Radar visualization of detector accuracy across LOW, MID, and HIGH text-density bins. \textbf{(a)} Spatial and frequency-based methods; \textbf{(b)} foundation model and VLM API methods. The pronounced area contraction from LOW (blue) to HIGH (red) illustrates the Text Density Curse: nearly all methods lose accuracy as text density increases, with spatial/frequency detectors collapsing most severely.}
\label{fig:text_density_radar}
\end{figure}

\paragraph{Observation \& Insight.}
Table~\ref{tab:density_bins} and Figures~\ref{fig:text_density_radar},~\ref{fig:text_density_trend} visualize these effects across bins and percentiles. Most spatial and frequency-based detectors exhibit substantial degradation as text density increases: Gram-Net ($\Delta = -43.0\%$), CNNDetection ($\Delta = -41.1\%$), DFFreq ($\Delta = -37.8\%$), and ResNet50 ($\Delta = -37.3\%$) all lose more than 37\% of accuracy from LOW to HIGH density. Much of this collapse occurs at the LOW-to-MID transition (e.g., Gram-Net: 88.6\% $\to$ 45.2\%), suggesting that even moderate text volume disables spectral artifact detection. SPAI shows a comparatively modest decline ($\Delta = -22.9\%$), consistent with its partial reliance on semantic features. VLM APIs degrade more moderately (GPT-5.4: $-18.0\%$; Claude-Sonnet-4.6: $-11.3\%$), indicating that semantic reasoning partially but not fully offsets the density effect.

\textbf{Diagnostic insight.}
The degradation is consistent across all spatial and frequency detectors; GAPL is the sole exception ($\Delta = +3.8\%$), suggesting its foundation-model representations actively leverage textual structure rather than being disrupted by it.

\subsection{Text as Forensic Camouflage: Scene-Level Detection Gap}
\label{subsec:text_camouflage}

\paragraph{Setup \& Metric.}
To isolate the effect of text content itself, rather than density gradients, we compare Fake Accuracy on two GPT Image-2 fake sets: TextFake (text-rich) and an MSCOCO-based set (natural scenes, no visible text). The generator is fixed; scene type is the sole variable.
\paragraph{Observation \& Insight.}
Table~\ref{tab:text_vs_mscoco} shows a pervasive gap: most detectors score only 10--35\% Fake Acc.\ on TextFake but climb to 40--95\% on MSCOCO, with the generator held fixed. Among spatial/frequency methods, FreqNet and SAFE suffer the largest drops ($\Delta = +63.5\%$ and $+58.8\%$), confirming that dense glyph edges overwhelm the spectral cues they rely on; SPAI is the exception, with a smaller gap ($\Delta = +15.9\%$) and the highest TextFake Fake Acc.\ in its group (45.4\%). Foundation methods split into two patterns: CO-SPY and LaDeDa recover strongly on MSCOCO ($\Delta > 50\%$), whereas AIDE and DDA remain near-chance on both datasets ($\Delta < 13\%$), indicating generator-specific insensitivity independent of text. GAPL maintains the best performance across both domains ($\Delta = +19.5\%$). VLM APIs show moderate but consistent gaps; GPT-5.4 drops most sharply ($\Delta = +55.5\%$) while Gemini-3-Pro-Preview is the most resilient ($\Delta = +18.2\%$).

\textbf{Diagnostic insight.}
High $\Delta$ indicates that text actively suppresses otherwise functional features; low $\Delta$ with low absolute performance signals generator-specific failure. Only GAPL and Gemini achieve low $\Delta$ from a position of genuine TextFake competence.

\begin{table}[!ht]
\centering
\caption{Fake Accuracy (\%) on TextFake vs.\ MSCOCO, both from GPT Image-2. $\Delta = \text{MSCOCO} - \text{TextFake}$.}
\label{tab:text_vs_mscoco}
\resizebox{\columnwidth}{!}{
\begin{tabular}{llccc}
\toprule
\textbf{Category} & \textbf{Method} & \textbf{TextFake} & \textbf{MSCOCO} & $\boldsymbol{\Delta}$ \\
\midrule
\multirow{8}{*}{\shortstack[l]{Spatial \& \\ Frequency}}
& ResNet50~\cite{resnet50}       & 25.4 & 53.7 & $+$28.3 \\
& NPR~\cite{npr}            & 24.7 & 67.0 & $+$42.3 \\
& CNNDetection~\cite{cnndetect}   & 10.1 & 26.0 & $+$15.9 \\
& Gram-Net~\cite{gramnet}       &  9.6 & 40.3 & $+$30.7 \\
& DFFreq~\cite{dffreq}         & 32.5 & 77.3 & $+$44.8 \\
& FreqNet~\cite{freenet}        & 26.8 & 90.3 & \textbf{$+$63.5} \\
& SAFE~\cite{safe}           & 17.5 & 76.3 & \textbf{$+$58.8} \\
& SPAI~\cite{spai}           & 45.4 & 61.3 & $+$15.9 \\
\midrule
\multirow{7}{*}{Foundation}
& CLIPDetection~\cite{clipdetection}              & 52.0 & 83.7 & $+$31.7 \\
& LaDeDa~\cite{ladeda}                     & 30.5 & 82.7 & \textbf{$+$52.2} \\
& AIDE~\cite{aide}                       &  6.8 & 19.0 & $+$12.2 \\
& CO-SPY\textsubscript{ProGAN}~\cite{cospy} & 28.1 & 96.3 & \textbf{$+$68.2} \\
& CO-SPY\textsubscript{SD1.4}~\cite{cospy}  & 28.3 & 83.3 & \textbf{$+$55.0} \\
& DDA~\cite{dda}                        & 11.6 & 16.7 & $+$5.1  \\
& GAPL~\cite{gapl}                       & 76.2 & 95.7 & $+$19.5 \\
\midrule
\multirow{3}{*}{VLM API}
& Gemini-3-Pro-Preview~\cite{gemini3} & 55.1 & 73.3 & $+$18.2 \\
& GPT-5.4~\cite{gpt54}              & 37.8 & 93.3 & \textbf{$+$55.5} \\
& Claude-Sonnet-4.6~\cite{claudesonnet46}    & 50.9 & 87.3 & $+$36.4 \\
\bottomrule
\end{tabular}
}
\end{table}

\FloatBarrier
\begin{table*}[!t]
\centering
\caption{Entity OCR Hit Rate and detection accuracy (\%) per generator subset for representative methods, sorted by hit rate (ascending). Higher hit rate indicates stronger text rendering fidelity; GPT-Image-2, the highest-fidelity generator, achieves the lowest detection accuracy across all methods.}
\label{tab:entity_hit}
\resizebox{\textwidth}{!}{
\begin{tabular}{lc|ccc|ccc}
\toprule
\textbf{Generator} & \textbf{Hit Rate $\uparrow$} & \begin{tabular}{c}\textbf{GAPL}\\{\footnotesize \cite{gapl}}\end{tabular} & \begin{tabular}{c}\textbf{SPAI}\\{\footnotesize \cite{spai}}\end{tabular} & \begin{tabular}{c}\textbf{DFFreq}\\{\footnotesize \cite{dffreq}}\end{tabular} & \begin{tabular}{c}\textbf{Gemini}\\{\footnotesize \cite{gemini3}}\end{tabular} & \begin{tabular}{c}\textbf{Claude}\\{\footnotesize \cite{claudesonnet46}}\end{tabular} & \begin{tabular}{c}\textbf{GPT-5.4}\\{\footnotesize \cite{gpt54}}\end{tabular} \\
\midrule
Qwen-Image-2~\cite{qwenimage2}            & 41.8          & \textbf{86.5} & \textbf{89.8} & 62.8 & \textbf{94.9} & \textbf{95.2} & 89.3 \\
Nano-Banana-2~\cite{nanobanana2}           & 57.4          & 86.0 & 74.3 & \textbf{79.2} & 90.5 & 84.1 & 77.8 \\
Seedream-5-Lite~\cite{seedream5lite}         & 62.8          & 85.7 & 72.4 & 50.2 & 94.5 & 92.4 & \textbf{90.8} \\
GPT-Image-2 (TextFake)~\cite{gptimage2}  & \textbf{70.0} & 79.3 & 66.3 & 66.0 & 77.9 & 74.3 & 67.6 \\
\bottomrule
\end{tabular}
}
\end{table*}
\subsection{Cloaking via Rendering Fidelity}
\label{subsec:cloaking}

\paragraph{Setup \& Metric.}
To examine whether stronger text rendering makes generated images harder to detect, we define a diagnostic metric named \textbf{Entity OCR Hit Rate}. Conventional string-level metrics such as CER or WER are sensitive to layout variation, line breaks, and OCR alignment errors~\cite{ocr}. Our metric instead targets \textit{factual rendering fidelity}~\cite{ocr1}. For each generation prompt, we extract critical semantic anchors that should appear in the image, including percentages, monetary amounts, large numbers, drug dosages, stock symbols, and proper nouns. We then OCR the generated image and compute the exact recall rate of these entities in the OCR output~\cite{ocrbench}. This gives a targeted estimate of whether the generator has rendered the factual text that matters for visual plausibility.

\begin{equation}
\mathrm{HitRate}(x)=\frac{\#\{\mathrm{entities\ recovered\ by\ OCR}\}}{\#\{\mathrm{entities\ in\ prompt}\}}.
\end{equation}

\paragraph{Observation \& Insight.}
Table~\ref{tab:entity_hit} places the Entity OCR Hit Rate alongside detection accuracy for representative detectors across all four generator subsets, making the adversarial relationship directly visible. The rendering hierarchy: GPT-Image-2~\cite{gptimage2} (70.0\%) $>$ Seedream-5-Lite~\cite{seedream5lite} (62.8\%) $>$ Nano-Banana-2~\cite{nanobanana2} (57.4\%) $>$ Qwen-Image-2~\cite{qwenimage2} (41.8\%), maps onto a broadly inverse detection ordering.

\begin{table}[!ht]
\centering
\caption{Latin vs. non-Latin accuracy (\%).}
\label{tab:latin_nonlatin}
\resizebox{\columnwidth}{!}{
\begin{tabular}{lcccc}
\toprule
\textbf{Group} &
\begin{tabular}{c}\textbf{GAPL}\\{\footnotesize \cite{gapl}}\end{tabular} &
\begin{tabular}{c}\textbf{SPAI}\\{\footnotesize \cite{spai}}\end{tabular} &
\begin{tabular}{c}\textbf{DFFreq}\\{\footnotesize \cite{dffreq}}\end{tabular} &
\textbf{Mean} \\
\midrule
Latin & 79.59 & 63.23 & 62.49 & 68.44 \\
Non-Latin & 79.02 & 68.74 & 68.62 & 72.13 \\
$\Delta$ & -0.57 & +5.51 & +6.13 & +3.69 \\
\bottomrule
\end{tabular}
}
\end{table}

GPT-Image-2 achieves the highest text-rendering fidelity and remains the hardest to detect across all methods. VLM accuracy drops substantially on this subset compared to the lower-fidelity Qwen-Image-2 (e.g., Gemini drops by 17.0 pp, Claude by 20.9 pp). Specialized detectors also struggle: SPAI shows a sharp monotonic decline as fidelity increases, confirming that better character rendering suppresses exploitable spectral artifacts, whereas the foundation-model-based GAPL degrades more gradually.

Interestingly, Seedream-5-Lite presents an anomaly. It evades frequency-based detection (DFFreq drops to near-chance 50.2\%) but remains easily detectable by VLMs, likely due to lingering semantic or layout anomalies. This supports a multi-mechanism view: stronger generators first eliminate spectral artifacts, while semantic cues only vanish at peak fidelity (e.g., GPT-Image-2), causing VLM performance to drop sharply.

\textbf{Diagnostic insight.} Higher rendering fidelity consistently impairs detection accuracy. As generators perfect character-level statistics, both spectral and semantic artifacts vanish, effectively closing the forensic window for current detectors.


\subsection{Threshold Collapse in the Wild}
\label{subsec:threshold_collapse}

\paragraph{Setup \& Metric.}
Real-world text-rich forgeries rarely remain in their original file state. Screenshots and paper are commonly recompressed by platforms, re-captured from displays, saved across applications, or degraded by low-light sensors. We therefore stress-test three representative detectors (GAPL, SPAI, and DFFreq) under five realistic perturbations with severity standardized to level 3: JPEG compression, Gaussian noise, brightness shift, contrast shift, and screenshot moir\'{e}. We report overall accuracy, and for DFFreq we further report real accuracy and fake accuracy to expose class-specific threshold shifts.


\begin{table}[t]
\centering
\caption{Perturbation robustness: overall accuracy (\%).}
\label{tab:perturb_overall}
\resizebox{\columnwidth}{!}{
\begin{tabular}{lcccc}
\toprule
\textbf{Perturbation} &
\begin{tabular}{c}\textbf{GAPL}\\{\footnotesize \cite{gapl}}\end{tabular} &
\begin{tabular}{c}\textbf{SPAI}\\{\footnotesize \cite{spai}}\end{tabular} &
\begin{tabular}{c}\textbf{DFFreq}\\{\footnotesize \cite{dffreq}}\end{tabular} &
\textbf{Mean} \\
\midrule
Baseline & 79.33 & 66.34 & 66.03 & 70.57 \\
JPEG & 57.10 & 52.75 & 50.00 & 53.28 \\
Gaussian Noise & 51.85 & 46.00 & 50.40 & \textbf{49.42} \\
Brightness & 70.10 & 58.65 & 56.35 & 61.70 \\
Contrast & 71.05 & 58.30 & 55.00 & 61.45 \\
Screenshot moir\'{e} & 49.90 & 57.70 & 54.80 & 54.13 \\
\bottomrule
\end{tabular}
}
\end{table}

\paragraph{Observation \& Insight.}
Table \ref{tab:perturb_overall} shows that Gaussian noise, JPEG compression, and screenshot moir\'{e} cause the largest degradation, reducing the three-method mean to near-chance accuracy. More importantly, Table \ref{tab:perturb_classwise} shows that the failure is not a smooth robustness decline. DFFreq exhibits a binary shift: under JPEG compression, real accuracy reaches 100.0\% while fake accuracy falls to 0.0\%; under Gaussian noise, fake accuracy rises to 99.6\% while real accuracy collapses to 1.2\%.

\textbf{Diagnostic insight.}
The binary flip (JPEG pushing toward all-real with real accuracy 100\% and fake accuracy 0\%; Gaussian noise pushing toward all-fake with real accuracy 1.2\% and fake accuracy 99.6\%) shows that routine transmission perturbations cause complete threshold collapse in frequency-domain detectors.

\begin{table}[t]
\centering
\caption{Class-wise perturbation behavior for DFFreq (\%). Real Acc.\ = accuracy on real images; Fake Acc.\ = accuracy on fake images.}
\label{tab:perturb_classwise}
\small 
\begin{tabular*}{\columnwidth}{@{\extracolsep{\fill}}lcc}
\toprule
\textbf{Condition} & \textbf{Real Acc.} & \textbf{Fake Acc.} \\
\midrule
Baseline & 99.60 & 32.46 \\
JPEG & 100.0 & \textbf{0.0} \\
Gaussian Noise & \textbf{1.2} & 99.6 \\
Screenshot moir\'{e} & 19.3 & 90.3 \\
\bottomrule
\end{tabular*}
\end{table}

\subsection{Anti-Intuitive Biases: Language and Scene}
\label{subsec:language_scene_bias}

\paragraph{Linguistic Typology Bias.}
The 28-language breakdown yields a counter-intuitive pattern: non-Latin scripts are not consistently harder for current detectors. As shown in Table \ref{tab:latin_nonlatin}, frequency-sensitive methods improve on non-Latin languages, with SPAI gaining +5.51 pp and DFFreq gaining +6.13 pp over their Latin-script performance. GAPL remains nearly unchanged.

\textbf{Diagnostic insight.}
The pattern is counter-intuitive: frequency-sensitive detectors perform better on non-Latin scripts, suggesting that certain script topologies expose rendering inconsistencies more clearly than Latin text.

\paragraph{Scene Distribution Bias.}
Scene modality exposes a complementary bias. Specialized detectors generally perform better on screenshots than on paper, while VLM APIs show the opposite trend. SPAI drops from 70.20\% on screenshots to 50.43\% on paper, a nearly 20 pp gap.

\textbf{Diagnostic insight.}
Physical acquisition artifacts (paper texture, sensor noise, moir\'{e}) overlap with the spectral cues that frequency-domain detectors associate with synthesis, while VLM APIs benefit from richer semantic context in images of paper, a complementarity that points to the need for hybrid approaches.

\begin{table}[t]
\centering
\caption{Accuracy by scene modality (\%).}
\label{tab:scene_breakdown}
\resizebox{\columnwidth}{!}{
\begin{tabular}{lccc}
\toprule
\textbf{Method} & \textbf{Screenshot} & \textbf{Paper} & $\boldsymbol{\Delta}$ \\
\midrule
GAPL~\cite{gapl} & 80.06 & 76.16 & +3.90 \\
SPAI~\cite{spai} & 70.20 & 50.43 & \textbf{+19.77} \\
DFFreq~\cite{dffreq} & 66.70 & 62.31 & +4.40 \\
Gemini-3-Pro-Preview~\cite{gemini3} & 75.32 & 88.07 & -12.75 \\
Claude-Sonnet-4.6~\cite{claudesonnet46}    & 72.10 & 83.74 & -11.64 \\
GPT-5.4~\cite{gpt54}              & 64.74 & 77.67 & -12.93 \\
\bottomrule
\end{tabular}
}
\end{table}

\section{Conclusion}
\label{sec:discussion}
We introduced TextFake, a 20,000-image benchmark that exposes a critical blind spot in AI-generated image detection: methods achieving strong results on natural-scene benchmarks suffer severe degradation on text-rich forgeries, with no evaluated method exceeding 80\% accuracy. Our diagnostic analyses trace this gap to three compounding factors: dense glyph strokes and anti-aliased edges suppress the spectral anomalies that spatial and frequency detectors depend on; improving text-rendering fidelity progressively erases generative footprints; and routine transmission perturbations such as JPEG recompression or screenshot moir\'{e} can flip a detector's decision entirely. Among all evaluated methods, GAPL proves the most robust, suggesting that foundation-model representations are better equipped to capture character-level structure. These findings underscore the need for detectors purpose-built for the statistical and robustness demands of text-rich imagery.

\clearpage
\section{Limitations}
\label{sec:limitations}

TextFake prioritizes evaluation depth over exhaustive coverage. Generator selection targets current best-in-class text-rendering capability; the modular design supports incremental expansion as the field evolves, consistent with standard benchmark practice. Annotation depth is concentrated in high-resource languages to enable statistically reliable per-language analysis, with multilingual and domain extension left as directions for future community efforts. The zero-shot, binary-classification protocol reflects the most deployment-relevant evaluation regime; more specialized tasks such as source attribution and region-level analysis are natural extensions that build on this foundation.

\section{Ethics Statement}
\label{sec:ethics}

A portion of the real images depict identifiable individuals; access to the full dataset requires completing a data-use agreement that restricts redistribution and limits use to non-commercial research purposes. Synthetic images are released solely for evaluating AIGI detection methods; the agreement explicitly prohibits any use facilitating misinformation. Annotators provided informed consent and were compensated above local minimum wage.

\bibliography{refs}

\appendix
\begingroup
\setlength{\textfloatsep}{6pt plus 1pt minus 2pt}
\setlength{\floatsep}{4pt plus 1pt minus 2pt}
\setlength{\intextsep}{4pt plus 1pt minus 2pt}
\setlength{\abovecaptionskip}{2pt}
\setlength{\belowcaptionskip}{0pt}
\renewcommand{\arraystretch}{0.95}
\setlength{\tabcolsep}{3pt}

\section{Dataset Details}
\label{app:dataset_stats}

\begin{figure*}[!ht]
\centering
\includegraphics[width=\textwidth]{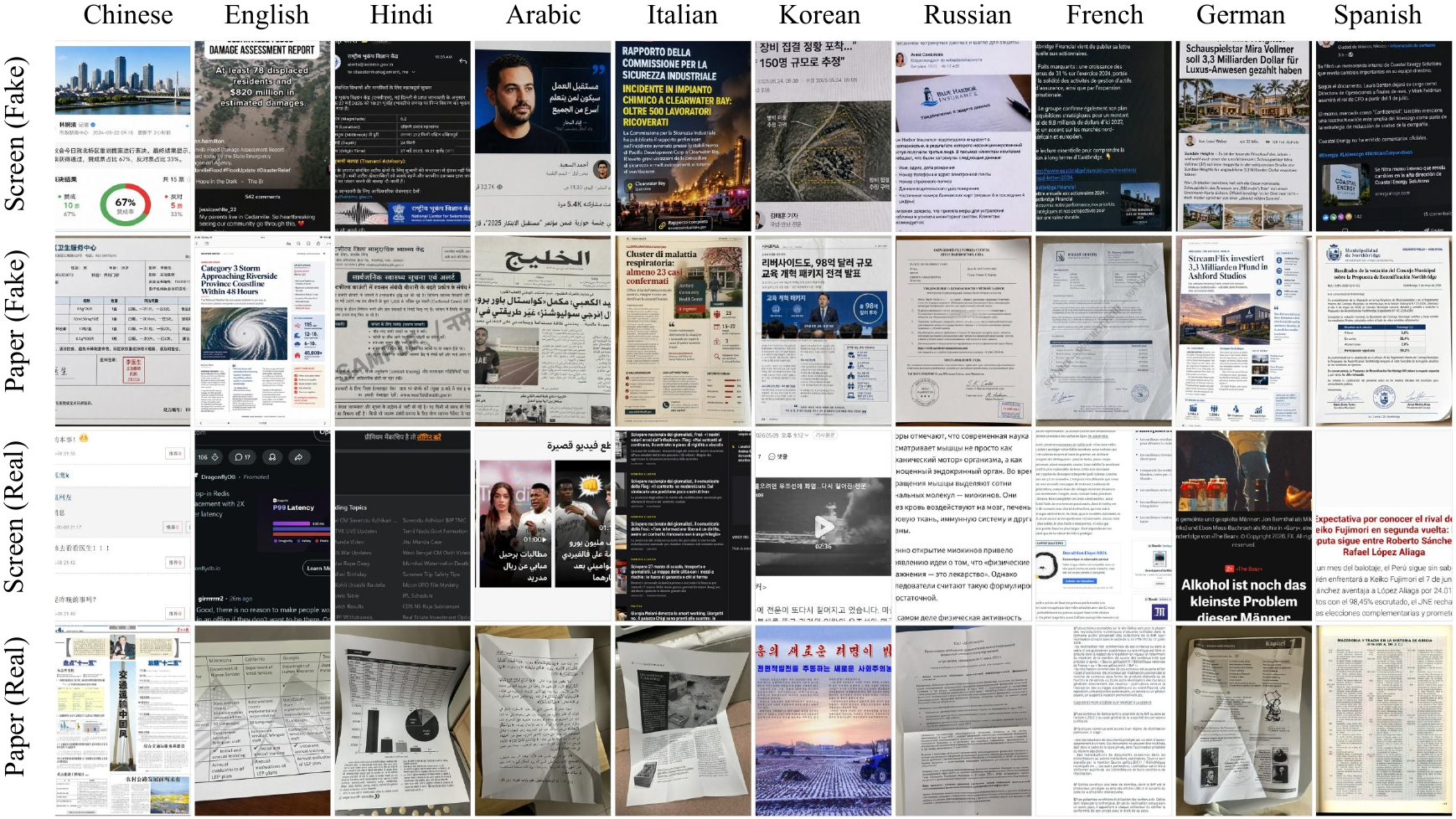}
\caption{Representative TextFake samples across ten languages and four authenticity--modality conditions. Rows (top to bottom): Screen~(Fake), Paper~(Fake), Screen~(Real), Paper~(Real). Columns: Chinese, English, Hindi, Arabic, Italian, Korean, Russian, French, German, Spanish.}
\label{fig:samples_app}
\end{figure*}

The \textbf{TextFake} dataset is available at \url{https://huggingface.co/datasets/Yuning0123/TextFake}. Figure~\ref{fig:samples_app} presents a structured sample grid from TextFake, organized into four rows and ten columns. Rows correspond to the four authenticity-modality conditions (Screen~Fake, Paper~Fake, Screen~Real, Paper~Real); columns cover ten representative languages (Chinese, English, Hindi, Arabic, Italian, Korean, Russian, French, German, Spanish). The grid illustrates the breadth of visual variation that detection methods must handle: glyph topologies range from Latin alphabets and CJK ideographs to Arabic cursive, Devanagari, and Hangul; layout conventions span social media feeds, newspaper front pages, and photographed manuscripts; acquisition conditions include clean screenshots, moir\'{e}-affected display re-captures, and low-light paper captures with visible sensor noise. Even human observers face genuine difficulty distinguishing real from fake.

Table~\ref{tab:language_distribution} lists the 28 languages sorted by sample count. The distribution is intentionally top-heavy: Chinese (15.5\%) and English (13.6\%) together account for nearly 30\% of the benchmark, reflecting their prevalence in real-world multilingual information ecosystems. A second tier of ten languages (Hindi, Arabic, Italian, Korean, Russian, French, German, Spanish, Japanese, Portuguese) each contribute 3--6\%, while the remaining 16 languages (including Burmese, Amharic, Tagalog, and Hebrew) form a long tail at 1--3\% each. This distribution enables high-statistical-power evaluation on common languages while preserving coverage of typologically distinct scripts.

\begin{table*}[!ht]
\centering
\caption{Language distribution in TextFake (28 languages, sorted by count).}
\label{tab:language_distribution}
\footnotesize
\setlength{\tabcolsep}{3pt}
\begin{tabular*}{\textwidth}{@{\extracolsep{\fill}}lrrlrrlrrlrr}
\toprule
\textbf{Language} & \textbf{Count} & \textbf{\%} &
\textbf{Language} & \textbf{Count} & \textbf{\%} &
\textbf{Language} & \textbf{Count} & \textbf{\%} &
\textbf{Language} & \textbf{Count} & \textbf{\%} \\
\midrule
Chinese    & 3,108 & 15.5 & English    & 2,721 & 13.6 & Hindi      & 1,097 & 5.5 & Arabic    & 913 & 4.6 \\
Italian    &   895 &  4.5 & Korean     &   858 &  4.3 & Russian    &   846 & 4.2 & French    & 793 & 4.0 \\
German     &   791 &  4.0 & Spanish    &   722 &  3.6 & Japanese   &   710 & 3.6 & Portuguese& 614 & 3.1 \\
Burmese    &   613 &  3.1 & Indonesian &   501 &  2.5 & Amharic    &   483 & 2.4 & Polish    & 470 & 2.4 \\
Ukrainian  &   439 &  2.2 & Marathi    &   439 &  2.2 & Vietnamese &   416 & 2.1 & Dutch     & 397 & 2.0 \\
Urdu       &   379 &  1.9 & Turkish    &   375 &  1.9 & Thai       &   288 & 1.4 & Bengali   & 250 & 1.3 \\
Swahili    &   250 &  1.3 & Persian    &   250 &  1.3 & Tagalog    &   207 & 1.0 & Hebrew    & 175 & 0.9 \\
\bottomrule
\end{tabular*}
\end{table*}

Table~\ref{tab:source_distribution} summarizes the three source categories that make up the real subset. Real screenshots (8,002 images, 40\% of the full benchmark) are collected from live web sources. Real documents (1,998 images, 10\%) consist of photographed or scanned physical media, including printed materials and handwritten manuscripts. Fake synthetics (10,000 images, 50\%) are fully AI-generated using structured prompts sampled to match the real subset's scene, topic, and language distributions. The 80:20 screen-to-paper ratio is preserved in both real and fake subsets, ensuring that scene-modality imbalance does not function as an authenticity signal.

\begin{table}[!ht]
\centering
\caption{Dataset composition by source.}
\label{tab:source_distribution}
\footnotesize
\setlength{\tabcolsep}{3pt}
\begin{tabular}{lcc}
\toprule
\textbf{Source} & \textbf{Count} & \textbf{Share} \\
\midrule
Fake synthetics  & 10,000 & 50.0\% \\
Real screenshot  &  8,002 & 40.0\% \\
Real document    &  1,998 & 10.0\% \\
\bottomrule
\end{tabular}
\end{table}

\section{Real and Fake Accuracy Breakdown}
\label{app:realfake}

Table~\ref{tab:realfake_full} extends the aggregate accuracy results in Table~\ref{tab:main_results} with a per-class breakdown. Real~Acc.\ measures the fraction of authentic images correctly classified; Fake~Acc.\ (equivalently, TPR) measures the fraction of AI-generated images correctly identified as fake. These two metrics capture qualitatively different failure modes, and their divergence exposes the systematic biases that aggregate accuracy obscures.

Among specialized detectors, a strong conservative bias is apparent. DFFreq achieves Real~Acc.\ of 99.6\%, but its Fake~Acc.\ falls to only 32.5\%, little better than random for a balanced binary task. This pattern reflects decision thresholds calibrated on natural-image benchmarks where synthesis artifacts are more pronounced; on text-rich content, these detectors default to predicting authenticity. SPAI shows a milder version of the same bias (Real~Acc.\ 87.5\%, Fake~Acc.\ 45.4\%). GAPL stands apart as the only specialized detector with a roughly balanced profile: Real~Acc.\ 82.4\% versus Fake~Acc.\ 76.2\%, a gap of only 6.2 pp, suggesting that its foundation-model representations are less anchored to spectral artifacts that text disrupts.

VLM APIs exhibit an analogous asymmetry, though somewhat less extreme. Claude-Sonnet-4.6 achieves a near-perfect Real~Acc.\ of 99.0\%, yet its Fake~Acc.\ is only 50.9\%; GPT-5.4 pairs 97.3\% Real~Acc.\ with 37.8\% Fake~Acc. Gemini-3-Pro-Preview produces the most balanced profile among VLMs (Real~Acc.\ 93.5\%, Fake~Acc.\ 55.1\%), which explains its leading overall accuracy. The result suggests that even models with strong visual and language reasoning exhibit an implicit authenticity prior that over-predicts real labels on this domain.

\begin{table}[!ht]
\centering
\caption{Per-class accuracy (\%) on TextFake. Acc.\ = overall accuracy; Real Acc.\ = accuracy on real images; Fake Acc.\ = accuracy on fake images (true positive rate).}
\label{tab:realfake_full}
\scriptsize
\setlength{\tabcolsep}{2pt}
\begin{tabular}{llccc}
\toprule
\textbf{Category} & \textbf{Method} & \textbf{Acc.} & \textbf{Real Acc.} & \textbf{Fake Acc.} \\
\midrule
\multirow{3}{*}{Specialized}
& \begin{tabular}{@{}l@{}}GAPL\\{\footnotesize \cite{gapl}}\end{tabular}      & 79.3 & 82.4 & 76.2 \\
& \begin{tabular}{@{}l@{}}SPAI\\{\footnotesize \cite{spai}}\end{tabular}      & 66.4 & 87.5 & 45.4 \\
& \begin{tabular}{@{}l@{}}DFFreq\\{\footnotesize \cite{dffreq}}\end{tabular}    & 66.0 & 99.6 & 32.5 \\
\midrule
\multirow{3}{*}{VLM API}
& \begin{tabular}{@{}l@{}}Gemini-3-Pro-Preview\\{\footnotesize \cite{gemini3}}\end{tabular} & 78.1 & 93.5 & 55.1 \\
& \begin{tabular}{@{}l@{}}Claude-Sonnet-4.6\\{\footnotesize \cite{claudesonnet46}}\end{tabular}    & 74.4 & 99.0 & 50.9 \\
& \begin{tabular}{@{}l@{}}GPT-5.4\\{\footnotesize \cite{gpt54}}\end{tabular}              & 67.6 & 97.3 & 37.8 \\
\bottomrule
\end{tabular}
\end{table}

\section{Per-Language Detection Accuracy}
\label{app:perlang}

Table~\ref{tab:perlang_specialized} reports detection accuracy for each of the 28 languages under three specialized detectors. Languages are sorted by mean accuracy (descending) to highlight the performance gradient from most- to least-detectable scripts. The spread is striking: the gap between the highest-mean language (Burmese, 86.2\%) and the lowest (Swahili, 49.3\%) spans 37 percentage points, comparable to the gap between a strong and a random-chance detector.

\begin{table*}[!ht]
\centering
\caption{Per-language overall accuracy (\%) for three specialized detectors on TextFake. Languages sorted by mean accuracy (descending). Bold indicates the highest score per language.}
\label{tab:perlang_specialized}
\footnotesize
\setlength{\tabcolsep}{3pt}
\begin{tabular*}{\textwidth}{@{\extracolsep{\fill}}llcccc}
\toprule
\textbf{Language} & \textbf{Script} & \begin{tabular}{@{}c@{}}\textbf{GAPL}\\{\footnotesize \cite{gapl}}\end{tabular} & \begin{tabular}{@{}c@{}}\textbf{SPAI}\\{\footnotesize \cite{spai}}\end{tabular} & \begin{tabular}{@{}c@{}}\textbf{DFFreq}\\{\footnotesize \cite{dffreq}}\end{tabular} & \textbf{Mean} \\
\midrule
Burmese    & Burmese       & 79.3 & \textbf{90.9} & 90.1 & 86.8 \\
Polish     & Latin         & 84.7 & \textbf{88.5} & 84.7 & 86.0 \\
Italian    & Latin         & 85.7 & 83.1 & \textbf{86.7} & 85.2 \\
Marathi    & Devanagari    & 79.0 & \textbf{86.8} & 85.2 & 83.7 \\
Ukrainian  & Cyrillic      & 81.1 & \textbf{84.5} & 82.9 & 82.8 \\
Hindi      & Devanagari    & \textbf{81.9} & \textbf{81.9} & 81.0 & 81.6 \\
Amharic    & Ge'ez         & 66.1 & 86.5 & \textbf{87.2} & 79.9 \\
German     & Latin         & 81.3 & 75.5 & \textbf{82.2} & 79.7 \\
Urdu       & Arabic        & \textbf{89.7} & 77.6 & 69.9 & 79.1 \\
Dutch      & Latin         & \textbf{80.9} & 65.0 & 83.6 & 76.5 \\
French     & Latin         & \textbf{83.4} & 73.0 & 72.4 & 76.3 \\
Turkish    & Latin         & \textbf{88.5} & 69.6 & 65.9 & 74.7 \\
Korean     & CJK           & \textbf{78.4} & 67.7 & 76.8 & 74.3 \\
Hebrew     & Hebrew        & \textbf{84.6} & 67.4 & 68.0 & 73.3 \\
Tagalog    & Latin         & \textbf{73.0} & \textbf{73.9} & 71.0 & 72.6 \\
Japanese   & CJK           & \textbf{79.9} & 66.8 & 69.2 & 72.0 \\
Russian    & Cyrillic      & \textbf{78.0} & 65.1 & 64.5 & 69.2 \\
Indonesian & Latin         & \textbf{74.3} & 61.9 & 70.5 & 68.9 \\
Arabic     & Arabic        & \textbf{80.8} & 61.8 & 63.4 & 68.7 \\
Chinese    & CJK           & \textbf{78.9} & 59.1 & 56.1 & 64.7 \\
Persian    & Arabic        & \textbf{89.3} & 58.8 & 42.8 & 63.6 \\
Bengali    & Bengali       & \textbf{72.8} & 58.0 & 55.2 & 62.0 \\
Portuguese & Latin         & \textbf{69.2} & 53.8 & 62.4 & 61.8 \\
Spanish    & Latin         & \textbf{74.1} & 56.1 & 54.9 & 61.7 \\
Vietnamese & Latin         & \textbf{70.9} & 48.1 & 65.4 & 61.5 \\
English    & Latin         & \textbf{81.2} & 51.7 & 39.6 & 57.5 \\
Thai       & Thai          & 65.3 & 39.9 & \textbf{62.5} & 55.9 \\
Swahili    & Latin         & \textbf{66.8} & 47.4 & 40.0 & 51.4 \\
\bottomrule
\end{tabular*}
\end{table*}

Several patterns stand out. Burmese (86.8\%) benefits from its visually complex letterforms, whose dense high-frequency glyph structure amplifies rendering inconsistencies that spectral detectors can exploit; SPAI and DFFreq both exceed 90\% on this language alone. Polish (86.0\%) and Italian (85.2\%) rank second and third despite using the Latin script, likely because their denser diacritic inventories introduce more discriminative high-frequency structure than plainer Latin scripts like English. At the opposite end, English ranks 26th out of 28 languages (57.5\%): DFFreq drops to 39.6\%, below chance, suggesting that the large volume of English text-image data in generator training enables highly realistic rendering that suppresses low-level artifacts. Swahili (51.4\%) reaches the lowest mean, essentially random across all three methods, indicating that even a familiar Latin script cannot guarantee detectability when the generator's training distribution is sparse for that language.

Table~\ref{tab:perlang_vlm} reports per-language accuracy for the three VLM APIs. The ordering differs substantially from Table~\ref{tab:perlang_specialized}, reflecting different competencies. Gemini-3-Pro-Preview leads on morphologically rich or visually complex non-Latin scripts: Marathi (100.0\%), Turkish (92.3\%), Urdu (87.5\%), and Korean (87.5\%). Claude-Sonnet-4.6 performs best on European languages (Italian 92.7\%, Ukrainian 88.6\%, Polish 89.1\%, and German 85.0\%), consistent with its strong multilingual representation of European corpora. GPT-5.4 achieves the single highest VLM score on Amharic (93.9\%), a script family where its training distribution appears to provide unusually discriminative OCR-like rendering intuition. All three VLMs share a weakness on Arabic, Portuguese, and Chinese, where Fake Acc.\ tends to be low, suggesting that high-quality generation for these scripts is particularly challenging to detect semantically.

\begin{table*}[!ht]
\centering
\caption{Per-language overall accuracy (\%) for three VLM APIs on TextFake. Languages sorted by Gemini-3-Pro-Preview accuracy (descending). Bold indicates the highest score per language.}
\label{tab:perlang_vlm}
\footnotesize
\setlength{\tabcolsep}{3pt}
\begin{tabular*}{\textwidth}{@{\extracolsep{\fill}}llccc}
\toprule
\textbf{Language} & \textbf{Script} & \begin{tabular}{@{}c@{}}\textbf{GPT-5.4}\\{\footnotesize \cite{gpt54}}\end{tabular} & \begin{tabular}{@{}c@{}}\textbf{Claude-Sonnet-4.6}\\{\footnotesize \cite{claudesonnet46}}\end{tabular} & \begin{tabular}{@{}c@{}}\textbf{Gemini-3-Pro-Preview}\\{\footnotesize \cite{gemini3}}\end{tabular} \\
\midrule
Marathi    & Devanagari & 85.1 & 89.4 & \textbf{100.0} \\
Dutch      & Latin      & 83.8 & 82.8 & \textbf{91.7}  \\
Turkish    & Latin      & 66.7 & 76.7 & \textbf{92.3}  \\
Korean     & CJK        & 75.3 & 73.2 & \textbf{87.5}  \\
Hindi      & Devanagari & 72.2 & 75.9 & \textbf{86.1}  \\
Indonesian & Latin      & 80.9 & 87.2 & \textbf{86.7}  \\
Ukrainian  & Cyrillic   & 71.1 & \textbf{88.6} & 85.7   \\
Burmese    & Burmese    & 81.7 & \textbf{86.7} & 83.3   \\
Polish     & Latin      & 89.1 & \textbf{89.1} & 86.7   \\
Amharic    & Ge'ez      & \textbf{93.9} & 89.8 & 84.2  \\
Italian    & Latin      & 88.5 & \textbf{92.7} & 81.0   \\
Japanese   & CJK        & 75.0 & \textbf{84.7} & 78.3   \\
German     & Latin      & 79.0 & \textbf{85.0} & 77.8   \\
Russian    & Cyrillic   & 69.1 & \textbf{79.7} & 75.0   \\
English    & Latin      & 65.6 & 77.2 & \textbf{77.5}  \\
French     & Latin      & 63.2 & 73.3 & \textbf{75.0}  \\
Urdu       & Arabic     & 58.7 & 63.0 & \textbf{87.5}  \\
Vietnamese & Latin      & 63.0 & 56.4 & \textbf{71.4}  \\
Chinese    & CJK        & 56.4 & 66.1 & \textbf{76.3}  \\
Spanish    & Latin      & 56.3 & 70.3 & \textbf{77.3}  \\
Arabic     & Arabic     & 53.5 & 58.1 & \textbf{72.0}  \\
Portuguese & Latin      & 46.0 & 58.2 & \textbf{58.8}  \\
Thai       & Thai       & \textbf{61.3} & 50.0 & 54.6  \\
\bottomrule
\end{tabular*}
\end{table*}

\section{Topic-Level Detection Accuracy}
\label{app:topic}

TextFake images are annotated with one of four semantic topic categories: \textit{Politics \& Military} (government actions, elections, armed conflicts), \textit{Society \& Livelihood} (social welfare, health, education), \textit{Technology \& Finance} (tech industry, market data, financial reporting), and \textit{Culture \& Entertainment} (sports, arts, celebrity, lifestyle). Real and fake subsets are balanced across topics to within 1.3 percentage points, ensuring that topic distribution cannot serve as a proxy for authenticity.

Table~\ref{tab:topic_breakdown} shows that Politics \& Military is the most detectable category across both specialized detectors (mean 72.9\%) and VLM APIs (Gemini 76.7\%, Claude 76.9\%). This is likely due to stereotyped visual templates (standardized mastheads, prominent bylines, and branded color schemes), which simultaneously amplify synthesis artifacts and provide richer semantic anchors for VLMs. Culture \& Entertainment is notably the strongest topic for Gemini (83.6\%), a 15-point margin over Claude (68.9\%), suggesting that visually distinctive entertainment imagery aligns particularly well with Gemini's training. Technology \& Finance and Society \& Livelihood present the greatest challenge for specialized detectors, with means of approximately 69\% and 68\%, reflecting more varied layouts and higher generator rendering quality in these domains. Among VLMs, Claude-Sonnet-4.6 and Gemini-3-Pro-Preview maintain consistently strong performance across all four categories, while GPT-5.4 shows a larger variance (69.6\% on Politics vs.\ 64.6\% on Society), indicating less stable cross-topic generalization.

\section{VLM Detection Prompt Template}
\label{app:prompt}
\begin{table*}[t]
\centering
\caption{Detection accuracy (\%) by topic category. Methods: GAPL~\cite{gapl}, SPAI~\cite{spai}, DFFreq~\cite{dffreq}, GPT-5.4~\cite{gpt54}, Claude-Sonnet-4.6~\cite{claudesonnet46}, and Gemini-3-Pro-Preview~\cite{gemini3}.}
\label{tab:topic_breakdown}
\scriptsize
\setlength{\tabcolsep}{3pt}
\begin{tabular*}{\textwidth}{@{\extracolsep{\fill}}lcccccc}
\toprule
\textbf{Topic} & \textbf{GAPL} & \textbf{SPAI} & \textbf{DFFreq} & \textbf{GPT} & \textbf{Claude} & \textbf{Gemini} \\
\midrule
Politics \& Military    & \textbf{81.6} & \textbf{70.3} & 66.8 & 69.6 & 76.9 & 76.7 \\
Culture \& Entertainment& 81.1 & 69.0 & 66.5 & 65.7 & 68.9 & \textbf{83.6} \\
Technology \& Finance   & 79.4 & 61.1 & \textbf{65.8} & \textbf{69.0} & \textbf{75.6} & 77.6 \\
Society \& Livelihood   & 76.3 & 64.2 & 64.6 & 64.6 & 72.4 & 77.0 \\
\bottomrule
\end{tabular*}
\end{table*}
All three VLM APIs (Gemini-3-Pro-Preview, GPT-5.4, Claude-Sonnet-4.6) are queried in a zero-shot setting with a fixed prompt template (Figure~\ref{fig:vlm_prompt}) designed to elicit forensically relevant analysis rather than content-level plausibility judgments. A key design choice is the explicit redirection toward low-level visual evidence: rather than asking whether an image ``looks real,'' the prompt directs models to inspect character-level rendering quality, image artifact signatures, and layout consistency with authentic platform conventions. This is necessary because high-fidelity forgeries can be semantically plausible while still carrying detectable rendering defects at the glyph or artifact level.


The prompt instructs the model to judge each image along four dimensions, including text rendering fidelity, visual artifacts, layout authenticity, and semantic coherence, and respond with a single-word label (\textbf{REAL} or \textbf{FAKE}) followed by a one-sentence justification. No few-shot examples or chain-of-thought scaffolding are provided, keeping the evaluation strictly zero-shot.

\begin{figure}[!ht]
\centering
\fbox{
\begin{minipage}{0.94\columnwidth}
\small
\textbf{System message}\\[0.4em]
You are an expert in digital image forensics. Your task is to determine whether a provided image is real (an authentic photograph or screen capture) or AI-generated/fabricated.\\[0.8em]
\textbf{User message}\\[0.4em]
\textit{
Please analyze the image carefully. Pay attention to the following cues:\\[0.3em]
1. \textbf{Text rendering}: Are characters properly formed, consistently spaced, and free of distortion or hallucinated glyphs?\\
2. \textbf{Visual artifacts}: Are there unnatural blending boundaries, compression inconsistencies, or irregular high-frequency patterns?\\
3. \textbf{Layout authenticity}: Do interface elements, borders, and typography match those of genuine platforms or print media?\\
4. \textbf{Semantic coherence}: Does the content make logical sense for the claimed source or publication?\\[0.3em]
Based on your analysis, respond with exactly one word on the first line (\textbf{REAL} or \textbf{FAKE}), followed by a single sentence of justification.
}
\end{minipage}
}
\caption{Zero-shot VLM detection prompt used for Gemini-3-Pro-Preview, GPT-5.4, and Claude-Sonnet-4.6. Only the binary label (first line) is used for computing accuracy.}
\label{fig:vlm_prompt}
\end{figure}
\section{GPU and API Usage}
\label{app:prompt}
To draw main conclusions, we conducted experiments using about 40 GPU hours of RTX A6000 and about 100,000 API requests, covering both API-based generation and VLM API evaluations.
\endgroup

\end{document}